%% file: main.tex
\documentclass[letterpaper, 10 pt, journal, twoside]{IEEEtran}

\IEEEoverridecommandlockouts       

\usepackage[pdftex]{graphicx}
\usepackage[hidelinks]{hyperref}
\hypersetup{
	colorlinks=true,
	anchorcolor=black,
	citecolor=green,
	linkcolor=red,
	filecolor=cyan,      
	urlcolor=magenta
}
\usepackage[hyperpageref]{backref} 
\usepackage{times}
\usepackage{epsfig}
\usepackage{amsmath}
\usepackage{amssymb}
\usepackage{cite}
\usepackage{color}
\usepackage{url}

\usepackage[font=small]{caption}

\usepackage{subcaption}
\usepackage{xspace}

\newcommand{\etal}{\textit{et al}. }
\newcommand{\ie}{\textit{i}.\textit{e}}
\newcommand{\eg}{\textit{e}.\textit{g}}

\begin{document}
	
	\markboth{IEEE Robotics and Automation Letters. Preprint Version. Accepted 
		May, 2018}
	{Yang \MakeLowercase{\textit{et al.}}: Challenges in Monocular Visual 
	Odometry}
	
	\author{Nan Yang$^{1, 2,\ast}$, Rui Wang$^{1, 2,\ast}$, Xiang Gao$^{1}$ and 
	Daniel 
	Cremers$^{1, 2}$%
		\thanks{Manuscript received: February, 24, 2018; Revised May, 11, 2018; 
			Accepted May, 31, 2018.}%
		\thanks{This paper was recommended for publication by Editor Cyrill 
		Stachniss 
			upon evaluation of the Associate Editor and Reviewers' comments.}%
		\thanks{$^{1}$The authors are with the Chair for Computer Vision and 
		Artificial Intelligence, Department of Informatics, Technical 
		University of Munich, Germany. 
		\href{https://vision.in.tum.de/}{https://vision.in.tum.de/}}
		\thanks{$^{2}$The authors are with Artisense. 
		\href{https://www.artisense.ai/}{https://www.artisense.ai/}}
		\thanks{$^{\ast}$Nan Yang and Rui Wang contributed equally to this 
		paper.}%
		\thanks{Digital Object Identifier (DOI): see top of this page.}
	}
	\title{Challenges in Monocular Visual Odometry: Photometric 
		Calibration, Motion Bias and \\ Rolling Shutter Effect}
	
	\maketitle
	
	\input{Abstract}
	\input{Introduction}

	\input{RelatedWork}
	\input{Evaluation}
	\input{Conclusion}
	
	\bibliographystyle{IEEEtran}
	\bibliography{egbib.bib}

\end{document}

%% file: Abstract.tex
\begin{abstract}
Monocular visual odometry (VO) and simultaneous localization and mapping (SLAM) 
have seen tremendous improvements in accuracy,
robustness and efficiency, and have gained increasing popularity over recent 
years.
Nevertheless, not so many discussions have been carried out to reveal the 
influences
of three very influential yet easily overlooked
aspects: photometric calibration, motion bias and rolling shutter effect.
In this work, we evaluate these three
aspects quantitatively on the state of the art of direct, feature-based and semi-direct methods,
providing the community with useful practical knowledge both for better applying
existing methods and developing new algorithms of VO and SLAM.
Conclusions (some of which are counter-intuitive) are drawn with both technical 
and empirical
analyses to all of our experiments. 
Possible improvements on existing methods are directed or proposed, such as a sub-pixel
accuracy refinement of ORB-SLAM which boosts its performance.
\end{abstract}

%% file: Introduction.tex
\section{Introduction}
\label{sec:intro}
\IEEEPARstart{M}{odern} visual SLAM systems usually have two basic components: 
VO and global map 
optimization. While the VO component incrementally estimates camera poses and 
builds up a local map, small errors are accumulated and over time the estimated 
camera poses start to drift away from their actual positions. If a previously 
visited location is detected, the drift can be eliminated by global map 
optimization using techniques like loop closure with pose graph 
optimization or 
bundle adjustment. 
VO, commonly considered as the system front-end, fundamentally determines the 
overall 
performance of a SLAM system.
During the past few years, the VO community has seen significant progress in 
improving algorithm accuracy, robustness and efficiency~\cite{Irani99allabout, 
torr1999feature, ptam, lsdslam, orbslam, dtam, svo1, dso, svo2}. Efforts 
have been made for different VO formulations, \ie, direct vs. 
feature-based methods, dense/semi-dense alternating optimization vs. sparse 
joint optimization. However, apart from these high-level diversities, it is 
still not clear how the performance can be influenced by the following  
low-level aspects:

\textbf{(a) Photometric calibration}. Pixels corresponding to the same 3D 
point
may have different intensities across images due to camera optical vignetting, 
auto gain and exposure controls.

\textbf{(b) Motion bias}. Running a VO method on the same sequence 
forward and 
backward sometimes can result in significantly different performances.

\textbf{(c) Rolling shutter effect}. Exposing pixels within one 
image at different timestamps can produce distortions that may introduce 
non-trivial errors into VO systems.

These three aspects can greatly affect the VO performance, yet their 
influences have not been systematically discussed and evaluated. 
In this work, we perform systematic and quantitative evaluations on the three 
most popular formulations of VO, namely direct, feature-based and semi-direct 
methods. Since evaluating all existing methods is not realistic, we select the 
state of the art of each family, \ie., DSO~\cite{dso}, 
ORB-SLAM~\cite{orbslam} (with its loop closure and global bundle 
adjustment 
functionalities turned off) and SVO\cite{svo2} (we use the updated 
version SVO 
2.0). Our goal is to deliver practical insights for better applying existing 
methods and further designing new algorithms by giving insightful technical and 
empirical analyses to all of our experimental results. 
Our main contributions are summarized as follows: 

(1) While it has been shown in~\cite{dso,tummono} that photometric calibration 
can significantly improve the performance of direct methods, it is still 
unclear how or why it can influence other formulations. We complete this discussion 
by performing thorough evaluations on all the three selected methods, draw 
counter-intuitive conclusions and analyze the possible reasons. 

(2) Although motion bias was unveiled in~\cite{tummono}, the problem was not 
studied there at all. In this work we exhaustively discuss the problem, analyze 
the reasons and perform experiments that support our conclusions.

(3) In~\cite{dso} the rolling shutter effect was tackled partly and indirectly 
by simply mimicking the effect using random pixel shifting. In this work we 
carry out evaluations on dataset that provides both global and simulated 
rolling shutter sequences. Besides, we further evaluate the selected methods on 
modern industrial level cameras, which normally have rolling shutters
but extremely fast readout speed.

(4) In all the related experiments in~\cite{dso,tummono}, only direct and 
feature-based methods were considered. In this work we add the popular 
representative of semi-direct methods~\cite{svo1, svo2} to all our evaluations.

(5) We propose possible improvements of existing methods, \eg., a sub-pixel 
accuracy refined version of ORB-SLAM delivering boosted performance. 


%% file: RelatedWork.tex
\section{Related Work}
\label{sec:relw}
In this section we briefly introduce the principles of the three VO formulations
together with their respective selected representatives. Afterwards we list the
datasets used for our experiments.

\subsection{Direct Methods}
Direct methods use either all pixels (dense)~\cite{dtam}, pixels with 
sufficiently large intensity gradient (semi-dense)~\cite{lsdslam}, or sparsely 
selected pixels (sparse)~\cite{dso} and minimize a photometric error obtained 
by direct image alignment on the used pixels, based on the brightness constancy 
assumption. Camera 
poses and pixel depths are estimated by minimizing the photometric error using 
non-linear optimization algorithms. Since much image information can be used, 
direct methods are robust in low-texture scenes and can deliver relatively 
dense 3D reconstructions. Consequently, due to the direct image alignment 
formulation, direct methods are very sensitive to unmodeled artifacts such as 
rolling shutter effect, camera auto exposure and gain control. More crucially, 
the brightness constancy assumption does not always hold in practice, which 
drastically reduces the performance of direct methods in environments with 
rapid lighting change.

\begin{figure}
\vspace{1.0em}
	\captionsetup[sub]{font=small, justification=centering}
	\centering
	\begin{subfigure}[t]{0.225\textwidth}
		\hspace*{-4.0mm}
		\includegraphics[width=\textwidth]
		{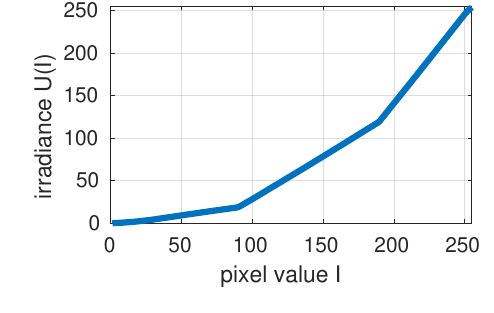}
		\vspace*{-2.0mm}
		\caption{Response with gamma correction.}\label{fig:phcexp1}
	\end{subfigure}
	\begin{subfigure}[t]{0.225\textwidth}
		\hspace*{-3.0mm}
		\includegraphics[width=\textwidth]
		{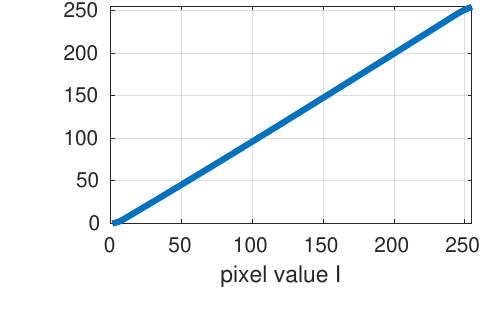}
		\vspace*{-2.0mm}
		\caption{Response without gamma correction.}\label{fig:phcexp2}
	\end{subfigure}
	\begin{subfigure}[t]{0.215\textwidth}
		\includegraphics[width=\textwidth]{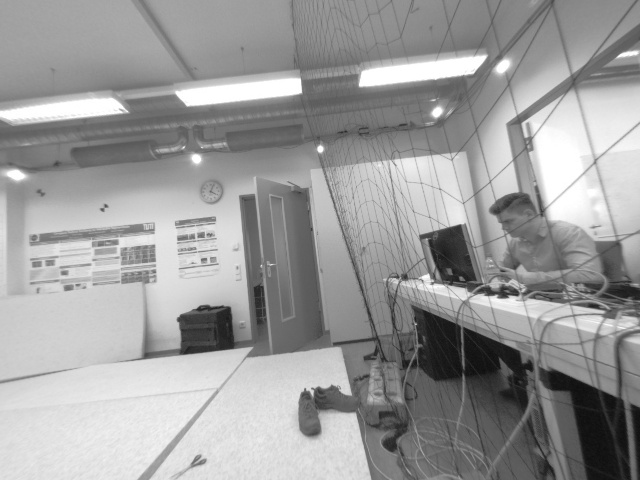}
		\caption{Original image.}\label{fig:phcexp3}
	\end{subfigure}
	\begin{subfigure}[t]{0.215\textwidth}
		\includegraphics[width=\textwidth]{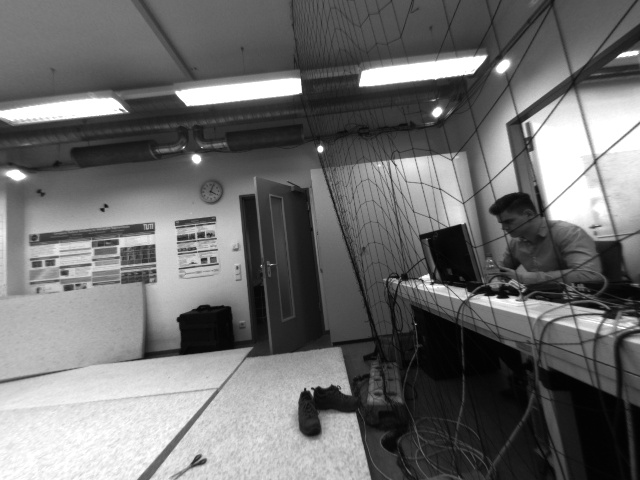}
		\caption{Calibrated image.}\label{fig:phcexp4}
	\end{subfigure}
	\caption{Example of photometric calibration. Camera response functions with
		gamma correction on (\subref{fig:phcexp1}) and off (\subref{fig:phcexp2}), 
		images before (\subref{fig:phcexp3}) and after (\subref{fig:phcexp4}) 
		photometric calibration.}
	\label{fig:phcexp}
	\vspace{-1.0em}
\end{figure}

\textbf{Direct Sparse Odometry (DSO)}.
DSO performs a novel sparse point sampling across image areas with sufficient 
intensity gradient. Reducing the amount of data enables real-time windowed 
bundle adjustment (BA). Obsolete and redundant information is 
marginalized with the Schur complement~\cite{slidingwindow}, and the First 
Estimate Jacobians technique is involved in the non-linear optimization 
process~\cite{huangfirst, slidingwindow} to avoid mathematical inconsistency. 
As a direct method, DSO is fundamentally based on the brightness constancy 
assumption, thus the authors proposed a photometric camera calibration pipeline 
to recover the irradiance images~\cite{dso,tummono}, which drastically 
increases the tracking accuracy~\cite{dso}. An example of photometric 
calibration is shown in Fig.~\ref{fig:phcexp}.

\vspace*{2mm}
\subsection{Feature-based Methods}
Feature-based methods extract a sparse set of key-features and match 
them across 
multiple frames. Camera poses and feature depths are 
estimated by minimizing the reprojection errors between feature pairs. As 
modern feature descriptors are to some extent invariant to illumination and 
view-point changes, feature-based methods are more robust than direct methods 
to brightness inconsistencies and large view-point changes. However, feature 
extraction and matching bring additional computational overhead, which limits 
the number of features that can be maintained in the system. The reconstructed 
3D maps therefore are much sparser and cannot be used directly for applications 
like obstacle avoidance and path planning. Moreover, in low-texture 
environments where not enough features can be extracted, tracking can easily 
get lost. 

\textbf{ORB-SLAM}. ORB-SLAM has become one of the most popular feature-based 
methods and has been widely adopted for a variety of applications. It uses ORB 
features~\cite{orbfeature} for all the tasks including tracking, mapping,
re-localization and loop closing. To track a new frame, motion-only BA is 
performed on its feature matches to estimate the initial pose, which is later 
refined by using all the feature matches in the local map and performing the 
pose optimization again. A covisibility graph is used to improve system 
efficiency by limiting the BA to a local covisible area. Unlike in DSO, in 
ORB-SLAM old points and keyframes are culled out directly from the active 
window without marginalization. To evaluate its VO performance, we disable its 
loop closure and global BA functionalities, and only focus on its tracking and 
local mapping components in this paper. 
 
\subsection{Semi-Direct Methods}

\textbf{Semi-Direct Visual Odometry (SVO)}. Semi-direct methods have been considered 
to be a hybrid of the two previously 
mentioned formulations. SVO~\cite{svo1} extracts Fast corners and perform 
direct image alignment on those areas for initial pose estimation. The 
feature 
extraction is later extended to also include edgelets~\cite{svo2} or line 
segments~\cite{gomez2016pl} to improve the robustness. The depths of the 
selected pixels are estimated from multiple observations using a recursive 
Bayesian depth filter. To reduce the drift caused by incremental estimations, 
poses and depths are refined by BA: Image patches from the reference frame and 
the current frame are aligned using an inverse compositional Lucas-Kanade 
algorithm, then the reprojection error is computed and minimized in BA using 
iSAM2~\cite{iSAM2}. Due to its exceptional high efficiency (around 400fps on a 
laptop) and low cost, SVO can be transplanted to devices with limited 
computational resources, thus has gained a high popularity in a wide range of 
robotics applications. 

\begin{figure*}[t]
	\centering
	\vspace{5mm}
	\begin{subfigure}[c]{0.65\textwidth}
		\includegraphics[width=\textwidth]{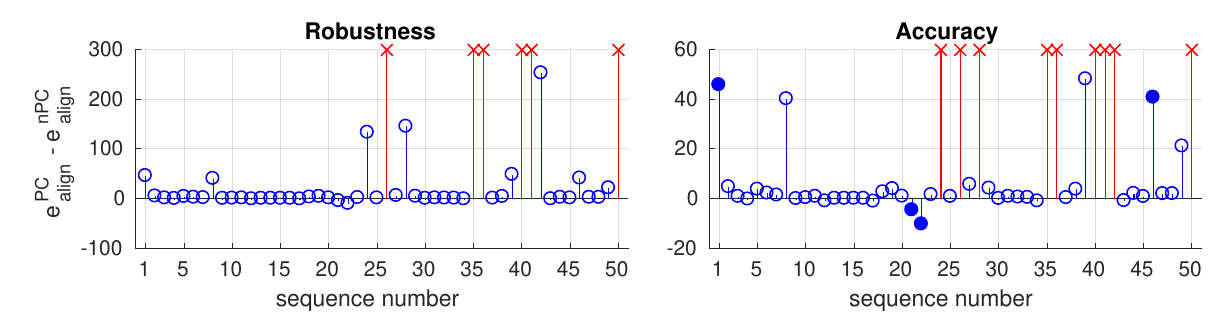}
	\end{subfigure}
	\begin{subfigure}[c]{0.3\textwidth}
		\label{fig:crfs2}
		\includegraphics[width=\textwidth]{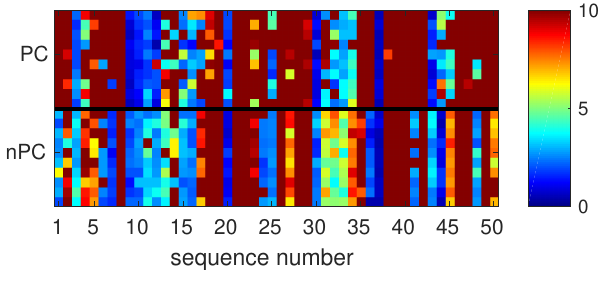}
	\end{subfigure}
	\vspace*{-2mm} 
	\caption{\textbf{Left}: Performance difference of ORB-SLAM on the TUM Mono Dataset. 
		The average of $e^{PC}_{align} - e^{nPC}_{align}$ is shown for each sequence, where $e^{PC}_{align}$
		and $e^{nPC}_{align}$ stand for the alignment error with and without photometric
		calibration, respectively. The 6 sequences on which tracking completely fails after photometric calibration are 
		marked in red. \textbf{Middle}: An enlarged view of the first plot. Sequences with performance
		differences larger than 50 are marked in red. Solid blue dots are used to mark the
		4 sequences shown in Fig.~\ref{fig:orb_matches}. \textbf{Right}: Color-encoded errors of 
		all runs. Data of ORB-SLAM without photometric calibration is obtained from~\cite{tummono}.}
	\label{fig:phccolorresult}
\end{figure*}

\subsection{Datasets}
The following datasets are used for our experiments which cover a variety of 
real-world settings, \eg, indoor/outdoor, texture/textureless, global/rolling 
shutters.

\textbf{The TUM Mono VO Dataset}~\cite{tummono} contains 50 sequences captured 
by a global shutter camera with two different lenses. Camera response function, 
dense attenuation factors and exposure time of each image are provided for 
photometric calibration. 

\textbf{The EuRoC MAC Dataset}~\cite{eurocmav} contains 11 sequences recorded 
by global shutter cameras mounted on a drone. Some of the sequences are quite 
challenging as they have extremely unstable motion and strong brightness change. 

\textbf{The ICL-NUIM Dataset}~\cite{iclnuim} has been extended by Kerl 
\etal~\cite{kerldataset} to provide both simulated rolling shutter and global 
shutter sequences of the same indoor environment. We use it for our experiments related 
the rolling shutter effect.

\textbf{The Cityscapes Dataset}~\cite{cityscapes} provides a long street view 
sequence captured by industrial rolling shutter cameras, which is 
used to evaluate how the selected methods work against realistic rolling 
shutter effect.


%% file: Evaluation.tex
\section{Evaluation}
\label{sec:evalres}

\subsection{Photometric Calibration}
\label{sec:photometric_calibration}
\begin{figure}
	\centering
	\includegraphics[width=\linewidth]{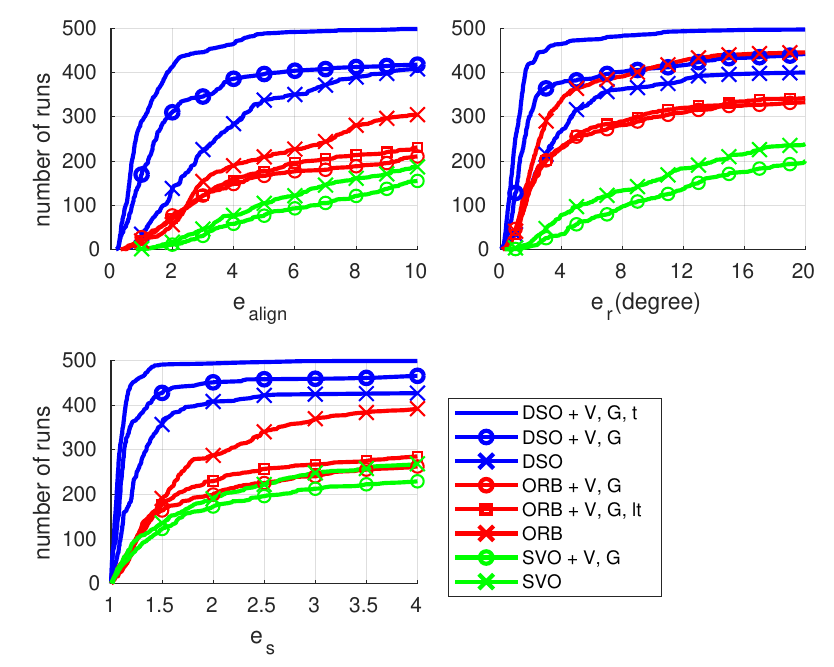}
	\vspace*{-7mm}
	\caption{Performance comparison of DSO, ORB-SLAM and SVO on sequences 
	with/without photometric calibration (camera response function $G$, 
	vignetting factors $V$). The alignment error $e_{align}$, rotation drift 
	$e_r$ (in degree) and scale drift $e_s$ are shown in the corresponding 
	subplots. ''lt'' stands for loosed thresholds for ORB feature extraction. 
	For reference, we also show results of DSO using camera exposure times $t$.}
	\label{fig:phcline}
	\vspace{-1.0em}
\end{figure}

In the first experiment, we evaluate the influence of photometric calibration on the 
selected methods, 
focusing more on analyzing its impacts on formulations other than direct method.
We use the 50 original sequences  from the TUM Mono VO Dataset and their 
corresponding ones after photometric calibration, \ie, with the nonlinear 
camera response function $G$ and pixel-wise  vignetting factors $V$ calibrated. 
Each method runs 10 times on each of these 100 sequences to account for  
non-deterministic behavior caused by \eg. multi-threading. The accumulative  
histogram (\ie, the number of runs that have errors less than the value given  
on the x axis) of the alignment error $e_{align}$\footnote{$e_{align}$ 
is the translational RMSE between the tracked trajectory when aligned 
to the start and the end segments of the ground truth trajectory. For 
details please refer to Eq.(14) of~\cite{tummono}.} (meter), rotation drift 
$e_r$ (degree) and scale drift $e_s$ are calculated~\cite{tummono} and shown 
in Fig.~\ref{fig:phcline}. It is worth noting that integrating the exposure 
times $t$ into the formulation of ORB-SLAM and SVO (not open-sourced) is not 
straightforward, therefore we do not use them for all three methods. For reference, we 
also show the results of DSO with all calibration information 
used, \ie, $G$, $V$ and $t$.

In this experiment, with $G$ and $V$ calibrated the performance of DSO 
increases significantly, which is not surprising as direct methods are built 
upon the brightness consistency assumption. Interestingly, photometric 
calibration reduces the overall performance of SVO, and for ORB-SLAM the 
performance decline is even larger. As both SVO and ORB-SLAM extract FAST 
corners, we suspect the feature extraction and feature matching of these 
methods are influenced by the photometric calibration.

\begin{figure*}
	\centering
	\vspace{10mm}
	\begin{subfigure}[b]{0.225\textwidth}
		\includegraphics[width=\textwidth]{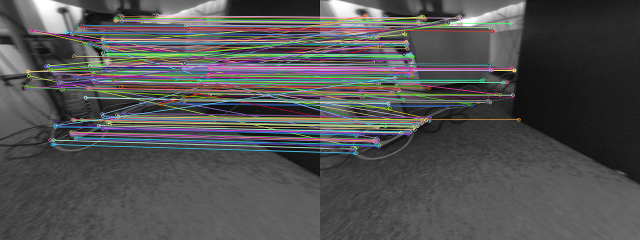}
	\end{subfigure}
	\begin{subfigure}[b]{0.225\textwidth}
		\includegraphics[width=\textwidth]{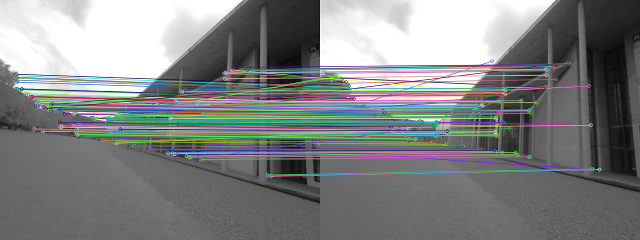}
	\end{subfigure}
	\begin{subfigure}[b]{0.225\textwidth}
		\includegraphics[width=\textwidth]{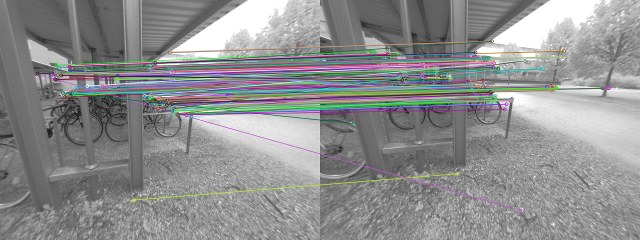}
	\end{subfigure}
	\begin{subfigure}[b]{0.225\textwidth}
		\includegraphics[width=\textwidth]{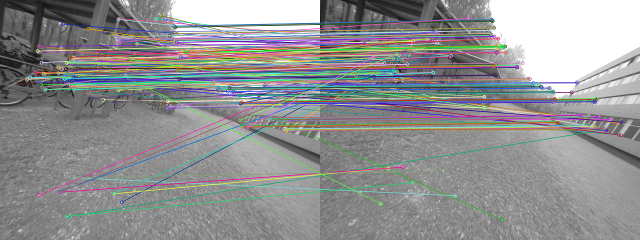}
	\end{subfigure}    \\
	
	\vspace*{-0.5mm}
	\begin{subfigure}[b]{0.225\textwidth}
		\includegraphics[clip, trim=1.82cm  0.6cm 1.335cm 0.5cm,width=\textwidth, height=0.5cm]{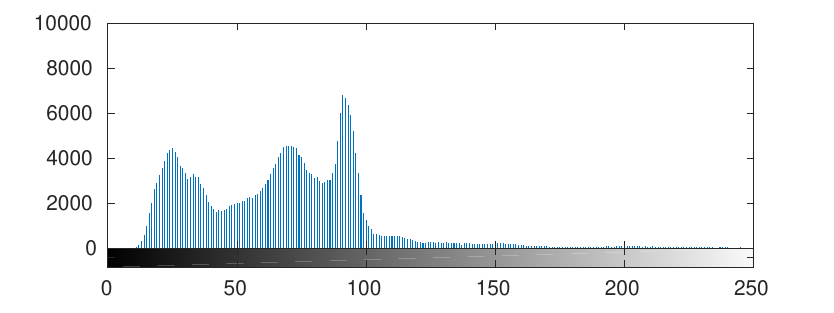}
		\vspace*{-6.8mm}
		\caption*{\textbf{\tiny{324 matches}}}
	\end{subfigure}
	\begin{subfigure}[b]{0.225\textwidth}
		\includegraphics[clip, trim=1.82cm  0.6cm 1.335cm 0.5cm,width=\textwidth, height=0.5cm]{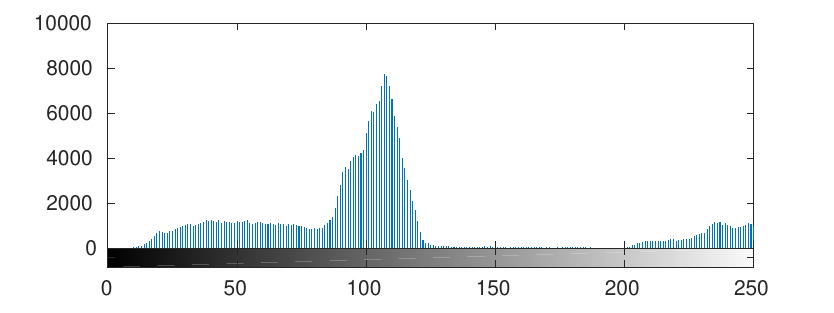}
		\vspace*{-6.8mm}
		\caption*{\textbf{\tiny{342 mathces}}}
	\end{subfigure}
	\begin{subfigure}[b]{0.225\textwidth}
		\includegraphics[clip, trim=1.82cm  0.6cm 1.335cm 0.5cm,width=\textwidth, height=0.5cm]{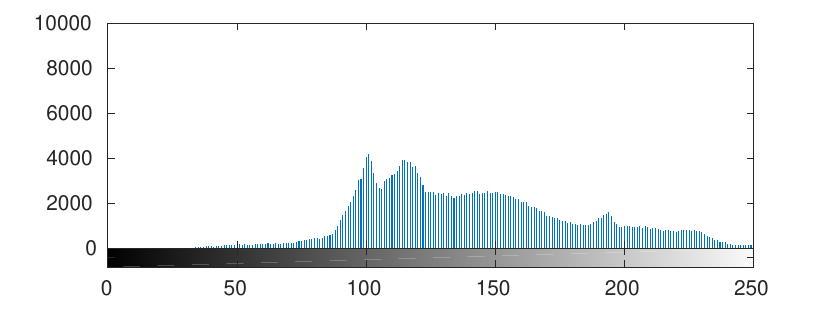}
		\vspace*{-6.8mm}
		\caption*{\tiny{206 matches}}
	\end{subfigure}
	\begin{subfigure}[b]{0.225\textwidth}
		\includegraphics[clip, trim=1.82cm  0.6cm 1.335cm 0.5cm,width=\textwidth, height=0.5cm]{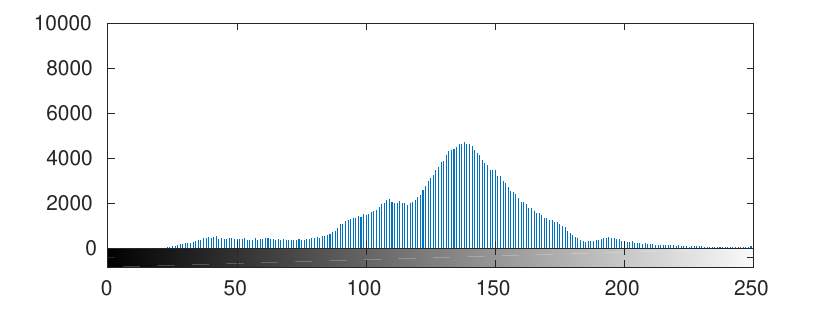}
		\vspace*{-6.8mm}
		\caption*{\tiny{278 matches}}
	\end{subfigure}    \\
	\begin{subfigure}[b]{0.225\textwidth}
		\includegraphics[width=\textwidth]{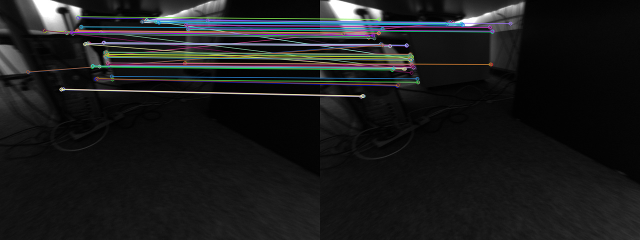}
	\end{subfigure}
	\begin{subfigure}[b]{0.225\textwidth}
		\includegraphics[width=\textwidth]{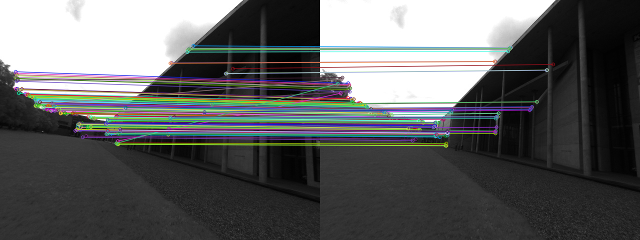}
	\end{subfigure}
	\begin{subfigure}[b]{0.225\textwidth}
		\includegraphics[width=\textwidth]{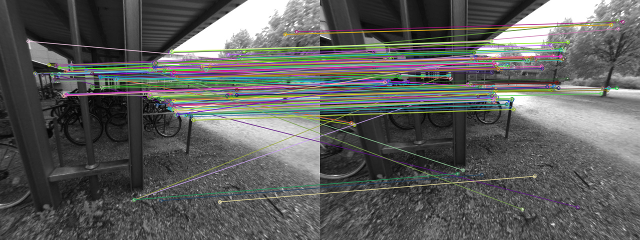}
	\end{subfigure}
	\begin{subfigure}[b]{0.225\textwidth}
		\includegraphics[width=\textwidth]{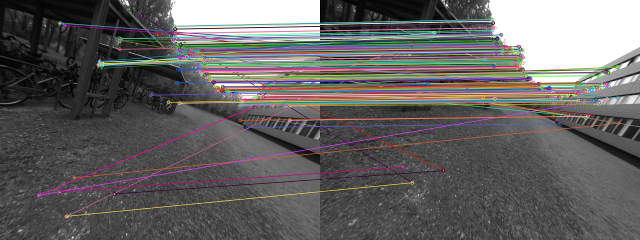}
	\end{subfigure}\\
	
	\vspace*{-0.5mm}
	\begin{subfigure}[b]{0.225\textwidth}
		\captionsetup{singlelinecheck = false, justification=centering}
		\includegraphics[clip, trim=1.82cm  0.6cm 1.335cm 0.5cm,width=\textwidth, height=0.5cm]{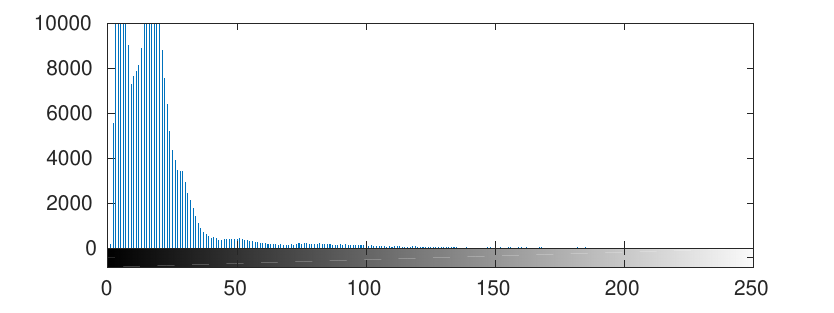}
		\vspace*{-6.8mm}
		\caption*{\tiny 86 matches \\ \small seq 01, frame 2085 }
	\end{subfigure}
	\begin{subfigure}[b]{0.225\textwidth}
		\captionsetup{singlelinecheck = false, justification=centering}
		\includegraphics[clip, trim=1.82cm  0.6cm 1.335cm 0.5cm,width=\textwidth, height=0.5cm]{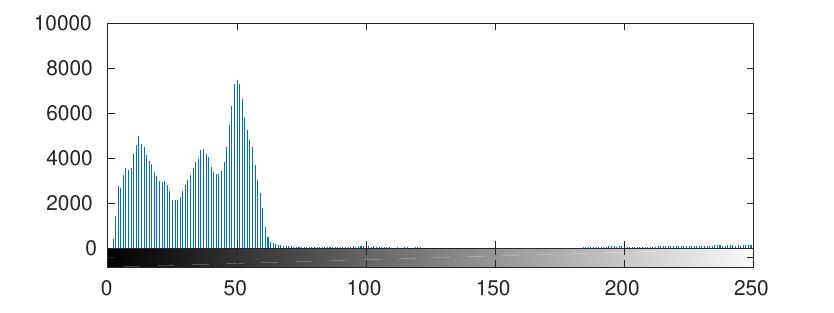}
		\vspace*{-6.8mm}
		\caption*{\tiny{257 mathces}\\ \small{seq 46, frame 2748}}
	\end{subfigure}
	\begin{subfigure}[b]{0.225\textwidth}
		\captionsetup{singlelinecheck = false, justification=centering}
		\includegraphics[clip, trim=1.82cm  0.6cm 1.335cm 0.5cm,width=\textwidth, height=0.5cm]{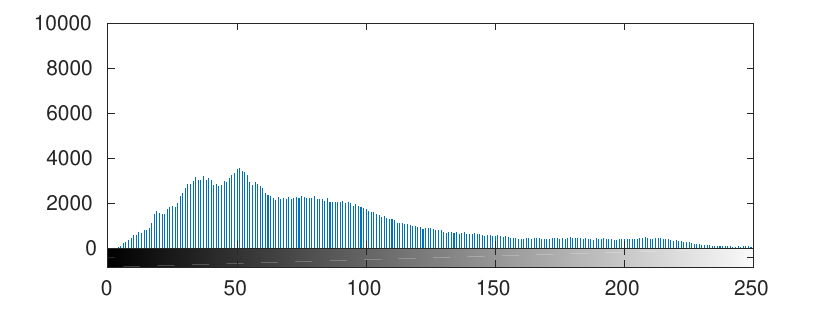}
		\vspace*{-6.8mm}
		\caption*{\textbf{\tiny{309 matches}}\\ \small{seq 21, frame 5369}}
	\end{subfigure}
	\begin{subfigure}[b]{0.225\textwidth}
		\captionsetup{singlelinecheck = false, justification=centering}
		\includegraphics[clip, trim=1.82cm  0.6cm 1.335cm 0.5cm,width=\textwidth, height=0.5cm]{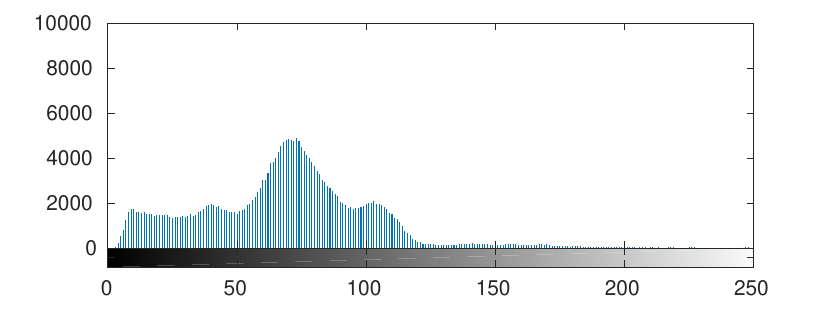}
		\vspace*{-6.8mm}
		\caption*{\textbf{\tiny{351 matches}}\\ \small{seq 22, frame 818}}
	\end{subfigure}    \\  
	\caption{Performance differences of ORB feature matching between 
	consecutive frames before and after photometric calibration. Top: before 
	calibration. Down: after calibration. Histogram of the left image is shown 
	under each image pair. As can be seen here, after photometric calibration 
	the numbers of matches decrease a lot on dark images, while increase 
	significantly on bright images.}
	\label{fig:orb_matches}
	\vspace{-0.5em}
\end{figure*}

\begin{figure*}
	\centering
	\begin{subfigure}[b]{0.15\textwidth}
		\includegraphics[clip, trim=0cm 0cm 0cm 2.5cm,width=\textwidth]{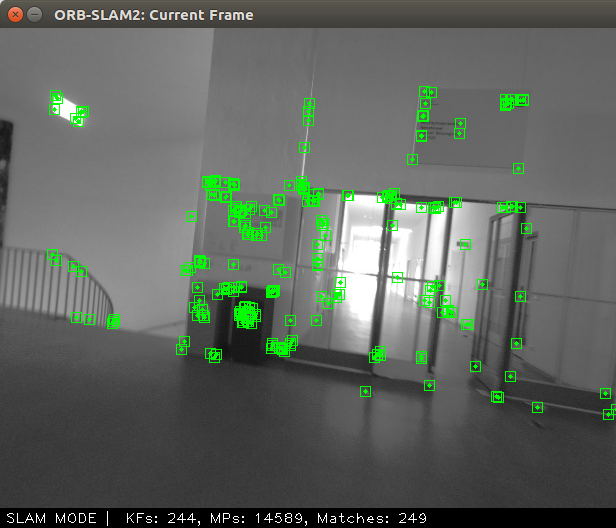}
		\vspace*{-6.8mm}
		\caption*{\tiny{\textbf{249} tracked}}
	\end{subfigure}
	\begin{subfigure}[b]{0.15\textwidth}
		\includegraphics[clip, trim=0cm 0cm 0cm 2cm,width=\textwidth]{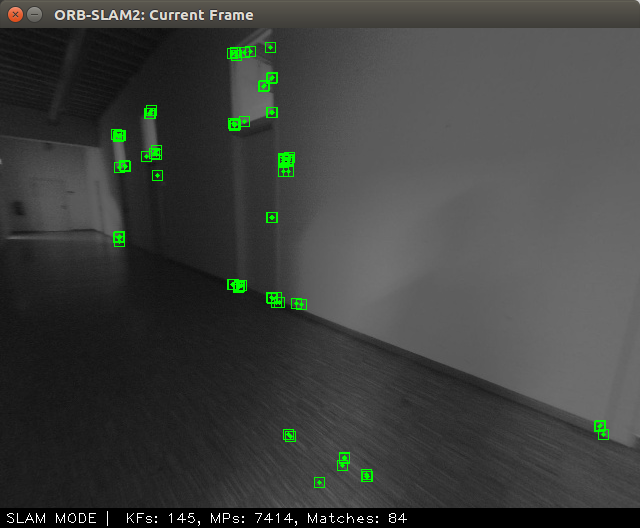}
		\vspace*{-6.8mm}
		\caption*{\tiny{\textbf{84} tracked}}
	\end{subfigure}
	\begin{subfigure}[b]{0.15\textwidth}
		\includegraphics[clip, trim=0cm 0cm 0cm 2cm,width=\textwidth]{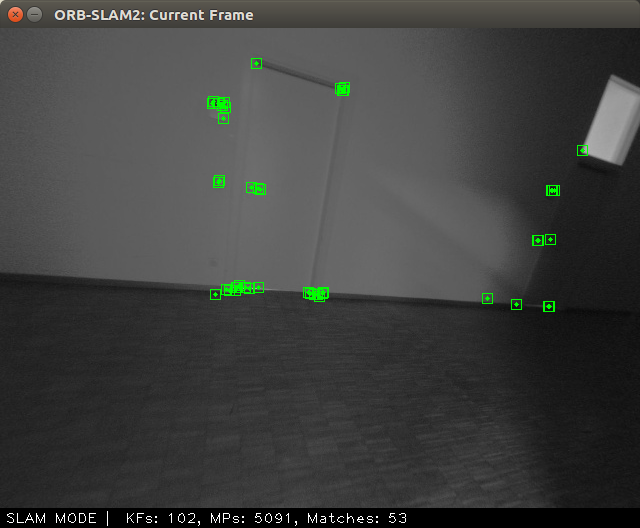}
		\vspace*{-6.8mm}
		\caption*{\tiny{\textbf{53} tracked}}
	\end{subfigure}
	\begin{subfigure}[b]{0.15\textwidth}
		\includegraphics[clip, trim=0cm 0cm 0cm 2cm,width=\textwidth]{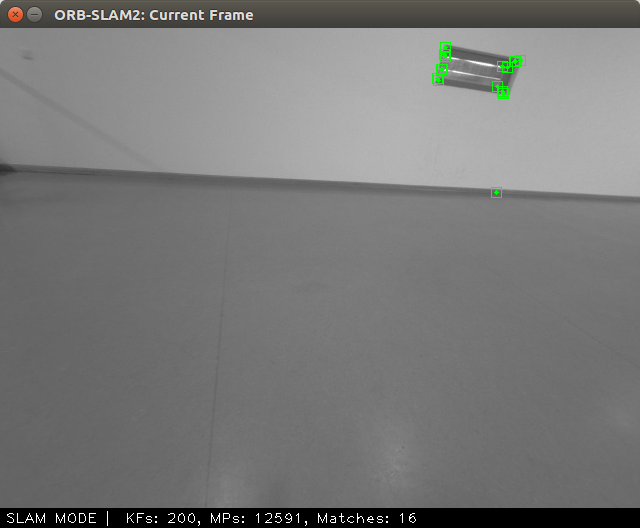}
		\vspace*{-6.8mm}
		\caption*{\tiny{\textbf{16} tracked}}
	\end{subfigure}
	\begin{subfigure}[b]{0.15\textwidth}
		\includegraphics[clip, trim=0cm 0cm 0cm 2cm,width=\textwidth]{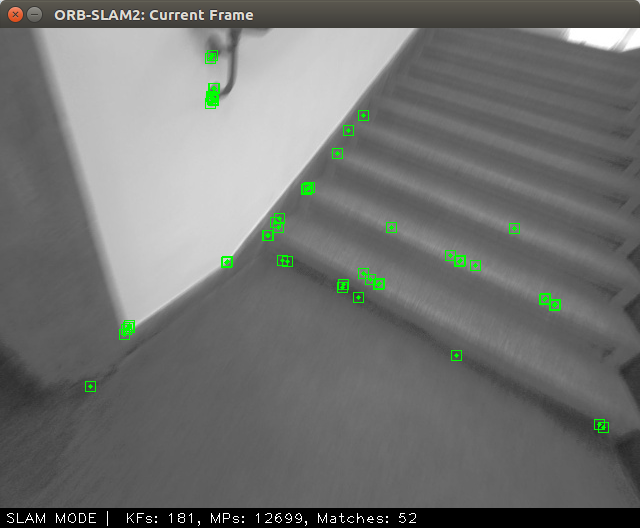}
		\vspace*{-6.8mm}
		\caption*{\tiny{\textbf{52} tracked}}
	\end{subfigure}
	\begin{subfigure}[b]{0.15\textwidth}
		\includegraphics[clip, trim=0cm 0cm 0cm 2cm,width=\textwidth]{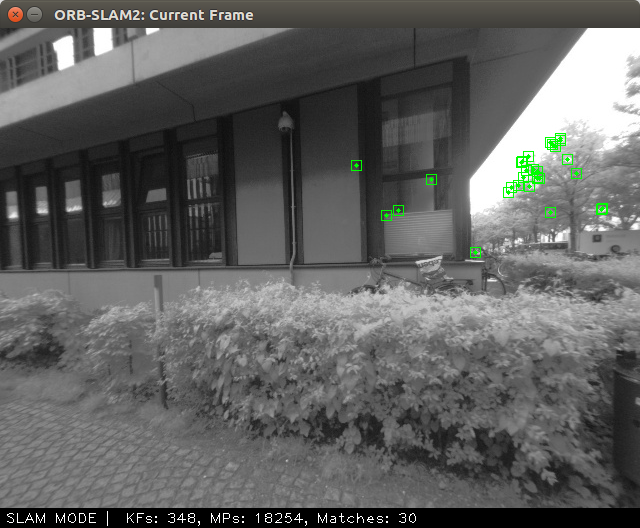}
		\vspace*{-6.8mm}
		\caption*{\tiny{\textbf{30} tracked}}
	\end{subfigure}    
	\begin{subfigure}[b]{0.15\textwidth}
		\includegraphics[clip, trim=0cm 0cm 0cm 2.5cm,width=\textwidth]{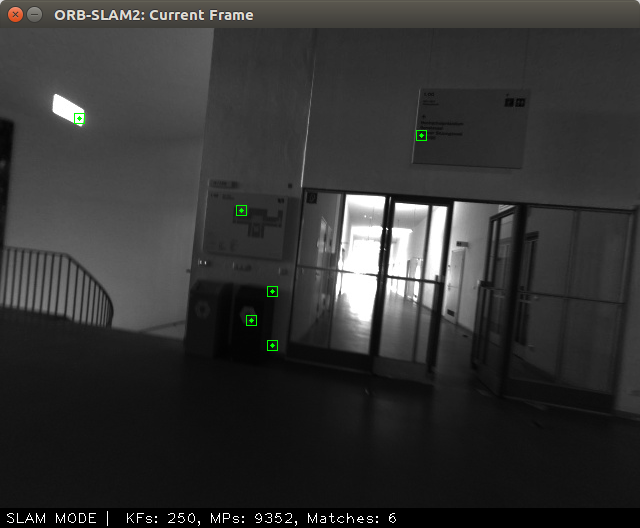}
		\vspace*{-6.8mm}
		\caption*{\tiny{6 tracked} \\ \small{seq 26}}
	\end{subfigure}
	\begin{subfigure}[b]{0.15\textwidth}
		\includegraphics[clip, trim=0cm 0cm 0cm 2cm,width=\textwidth]{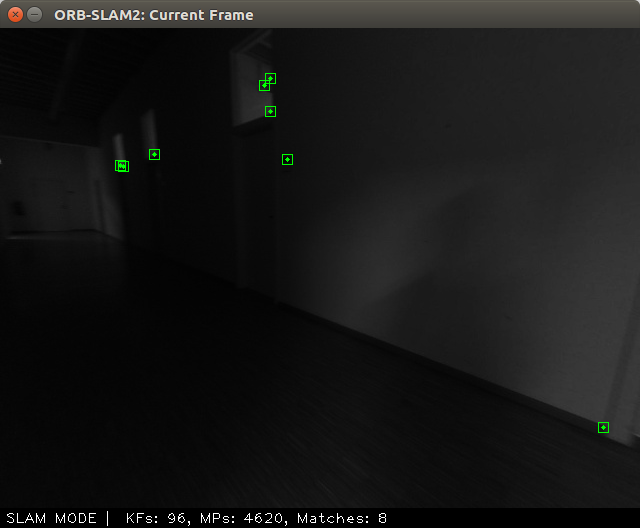}
		\vspace*{-6.8mm}
		\caption*{\tiny{8 tracked} \\ \small{seq 35}}
	\end{subfigure}
	\begin{subfigure}[b]{0.15\textwidth}
		\includegraphics[clip, trim=0cm 0cm 0cm 2cm,width=\textwidth]{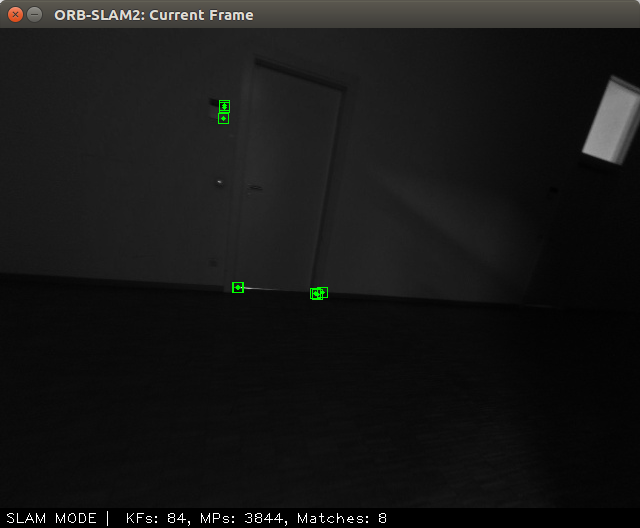}
		\vspace*{-6.8mm}
		\caption*{\tiny{8 tracked} \\ \small{seq 36}}
	\end{subfigure}
	\begin{subfigure}[b]{0.15\textwidth}
		\includegraphics[clip, trim=0cm 0cm 0cm 2cm,width=\textwidth]{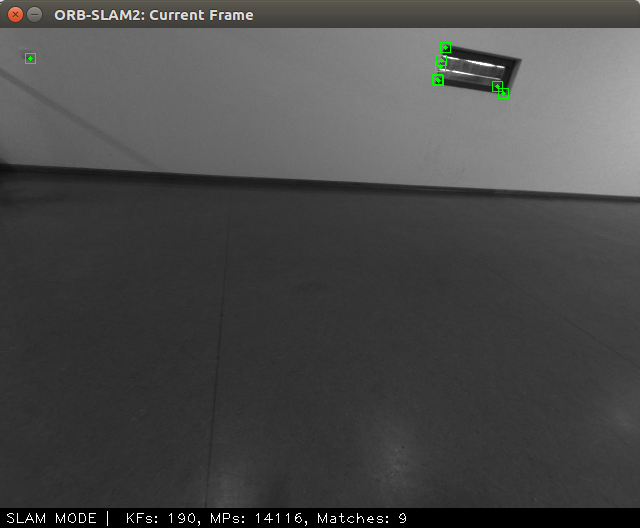}
		\vspace*{-6.8mm}
		\caption*{\tiny{9 tracked} \\ \small{seq 40}}
	\end{subfigure}
	\begin{subfigure}[b]{0.15\textwidth}
		\includegraphics[clip, trim=0cm 0cm 0cm 2cm,width=\textwidth]{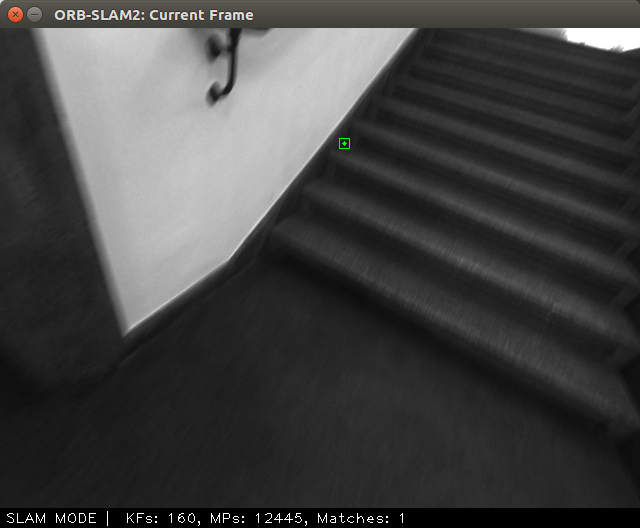}
		\vspace*{-6.8mm}
		\caption*{\tiny{1 tracked} \\ \small{seq 41}}
	\end{subfigure}
	\begin{subfigure}[b]{0.15\textwidth}
		\includegraphics[clip, trim=0cm 0cm 0cm 2cm,width=\textwidth]{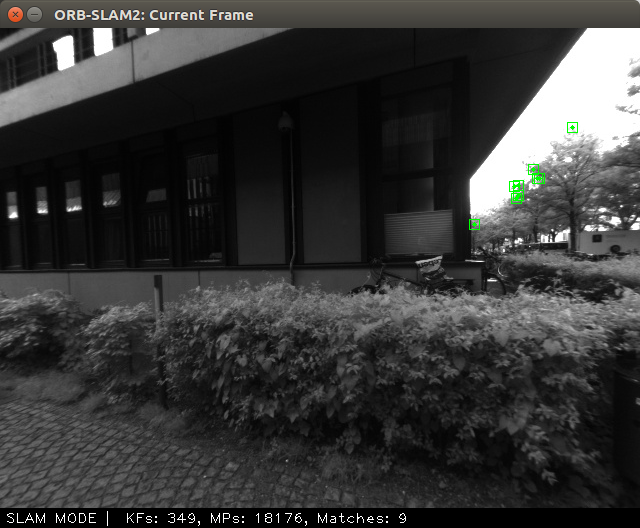}
		\vspace*{-6.8mm}
		\caption*{\tiny{9 tracked} \\ \small{seq 50}}
	\end{subfigure}
	\caption{Failure cases of the 6 sequences mentioned in the left of 
	Fig.~\ref{fig:phccolorresult} where ORB-SLAM lost its tracking. Top: before 
	photometric calibration. Down: after photometric calibration. Features 
	shown in the images are the ones projected from the active local map to the 
	current frame and successfully matched  there. After photometric 
	calibration, dark images become even darker and not enough features can be 
	matched on them.}
	\label{fig: tracklost}
	\vspace{-1.5em}
\end{figure*}

To better understand the results in Fig.~\ref{fig:phcline}, we further show the 
performance of ORB-SLAM on each sequence in Fig.~\ref{fig:phccolorresult}, 
where the differences of the alignment errors with/without photometric 
calibration $e^{PC}_{align} - e^{nPC}_{align}$ are shown in the left and 
middle, alignment errors of all runs are shown in the right. ORB-SLAM fails on 6 
sequences and generally performs worse on the other 
sequences. However, the performance decline is not consistent over sequences. 
By rechecking the inverse camera response function $G^{-1}$ in 
Fig.~\ref{fig:phcexp1}, we find the nonlinear function can be roughly 
divided into three linear parts with pixel values $I$ belonging to $[ 0, 90 )$, 
$[ 90, 190 )$ and $[ 190, 255 ]$. Due to the different slopes, applying 
$G^{-1}$ compresses intensities in $[ 0, 90 )$ and stretches the ones in $[ 
190, 255 ]$. In other words, it reduces contrast of dark areas while increases 
it for bright areas. As features like FAST and ORB generally work better on images with 
higher contrast (more evenly distributed intensity histogram), we further suspect that the 
performance declines of SVO and ORB-SLAM are mainly caused by dark frames.

To verify this another experiment is carried out: we extract ORB feature and 
match them on image pairs before and after being photometrically calibrated. 
Example results are shown in Fig.~\ref{fig:orb_matches}, where two of them are 
with dark images and the other two with bright ones. The numbers of feature 
matches and image histograms are shown under each image pair. As can be seen 
there, after photometric calibration the numbers of ORB feature matches 
decrease on the dark image pairs and increase on the bright ones.
Although sometimes the drop of the numbers may not seem 
crucial (\eg, the second column of Fig.~\ref{fig:orb_matches}), the effect can 
be accumulated over multiples frames. In Fig.~\ref{fig: tracklost} we show how 
the number of matches can drastically decrease when the system projects all 
features within the local map to the newest frame to search for correspondences.
As a result, only few features from the newest frame will be considered as 
inliers and added into the system, which is the main reason for the tracking 
failures in Fig.~\ref{fig:phccolorresult}.

We also try to loose the threshold for FAST extraction on the calibrated 
images. As shown 
in Fig.~\ref{fig:phcline}, although it slightly improves the  performance, it is still not 
comparable to the performance on the original  images. The reason is that feature 
extraction with lower threshold delivers more  noisy and unstable features. Moreover, as 
the images are internally represented by 8-bit unsigned integers, compressing the 
dynamic range of dark areas aggravates the discretization effect. The feature descriptors 
thus become less distinguishable, which corrupts the feature matching.


Recall that SVO extracts FAST corners and edgelets, the photometric calibration 
used in our experiments can reduce the number of successful extractions on dark 
image areas. On the other hand, instead of matching features, SVO matches image 
patches around those corner or edgelets, which is similar to direct image 
alignment and thus is less sensitive to the reduced intensity contrast. 
Moreover, SVO performs direct image alignment for initial pose estimation, 
to which photometric calibration is supposed to be beneficial. We believe these 
are the reasons for the reduced performance declines compared ORB-SLAM.

\subsection{Motion Bias}
\label{sec:motion_bias}
\begin{figure}[t]
\vspace{1.0em}
	\centering
	\begin{subfigure}[b]{0.43\textwidth}
		\includegraphics[width=\linewidth]{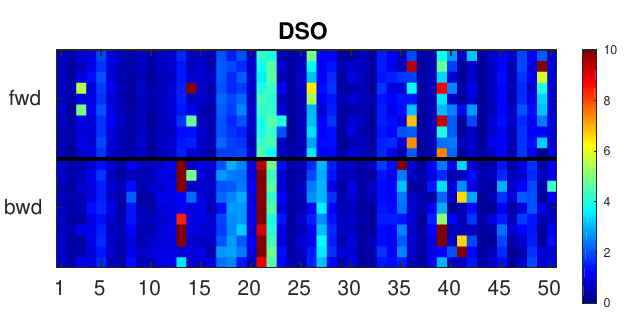}
		\vspace*{-4.8mm}
	\end{subfigure}
	\begin{subfigure}[b]{0.43\textwidth}
		\includegraphics[width=\linewidth]{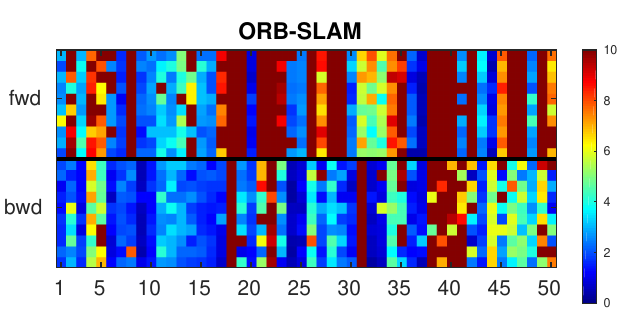}
		\vspace*{-4.8mm}
	\end{subfigure}
	\begin{subfigure}[b]{0.43\textwidth}
	\includegraphics[width=\linewidth]{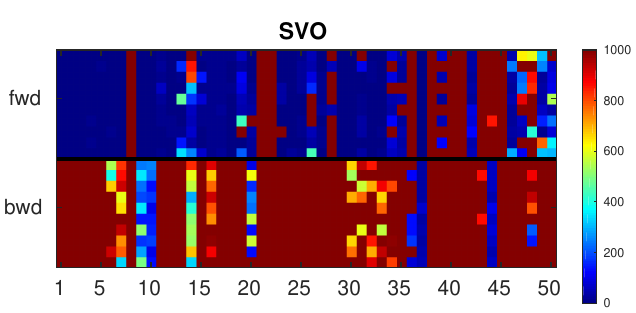}
	\end{subfigure}
	\caption{Results on TUM Mono VO Dataset running forward and backward. Each 
	method runs 10 times forward and 10 times backward on each sequence. The 
	alignment errors $e_{align}$ are color coded and shown as small blocks. For 
	the first two plots we use the results obtained in~\cite{dso}. Note that 
	for the results of SVO we use a different scale on the errors.}
	\label{fig:fwdbwd}
	\vspace{-1.0em}
\end{figure}

The term motion bias here refers to the 
difference of VO performance caused by running the same sequence forward and 
backward. Note that this is different from those studied 
in~\cite{dubbelman2009bias,dubbelman2012bias,farboud2014towards}. As 
shown in the top two plots in Fig.~\ref{fig:fwdbwd}, 
experiments in~\cite{dso,tummono} demonstrate that DSO does not suffer from 
such bias, but ORB-SLAM performs better when running backward. While this 
issue was raised there, no analysis, conclusion or possible remedies were given.
To get a more thorough understanding of motion bias, we first 
perform the same experiment for SVO and show the result in Fig.~\ref{fig:fwdbwd}. 
Surprisingly, SVO does not perform very well on this 
dataset and it gets very large alignment errors for all backward runs (note 
that we already use the settings recommended by one author of SVO 2.0). 
Generally speaking, the TUM Mono VO is a quite challenging dataset for 
monocular VO as it contains a lot of poorly textured areas. We suspect this is 
the main reason for the obtained results. However, as SVO 2.0 is not open-sourced, we 
cannot analyze further. We exclude SVO from the remaining 
experiments on the TUM Mono VO Dataset in this section.

Both for direct and feature-based methods, triangulation is a necessary step for
estimating depths of newly observed 3D points. Despite the cases with pure 
rotational camera motions, better depth estimation usually can be achieved with 
larger disparity between an image pair. When the camera is moving forward in a 
relatively open area, new points will emerge from the image center and have 
relatively small motions among consecutive frames. This pattern of optical flow 
introduces poorly initialized depths into the system. On the contrary, when 
moving backward, points close to the camera come into the field of view with 
large parallaxes, thus their depths are better initialized. We claim this is 
the main reason for the improved performance of ORB-SLAM when running backward. 

\begin{figure}
\vspace{1.0em}
	\centering
	\begin{subfigure}[b]{0.15\textwidth}
		\includegraphics[width=\textwidth]{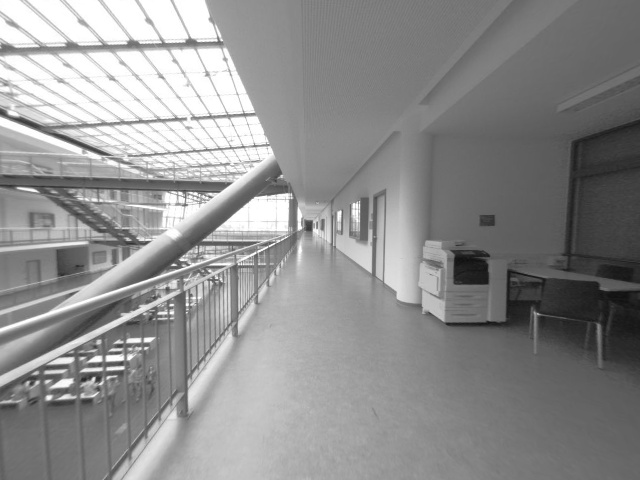}
		\vspace*{-6.8mm}
		\caption*{\tiny{seq 17}}
	\end{subfigure}
	\begin{subfigure}[b]{0.15\textwidth}
		\includegraphics[width=\textwidth]{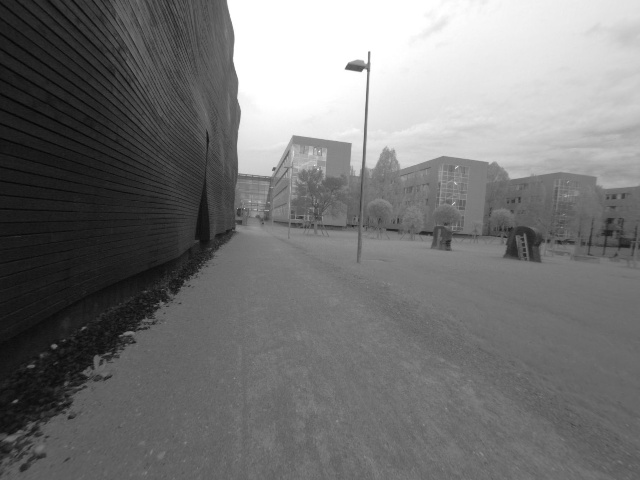}
		\vspace*{-6.8mm}
		\caption*{\tiny{seq 23}}
	\end{subfigure}
	\begin{subfigure}[b]{0.15\textwidth}
		\includegraphics[width=\textwidth]{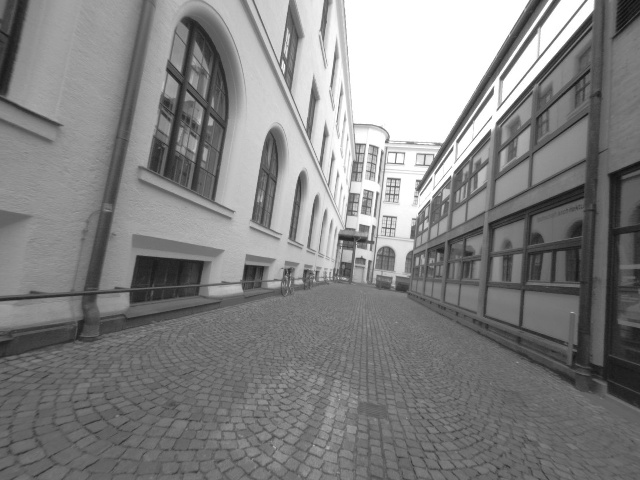}
		\vspace*{-6.8mm}
		\caption*{\tiny{seq 29}}
	\end{subfigure}
	\begin{subfigure}[b]{0.15\textwidth}
		\includegraphics[width=\textwidth]{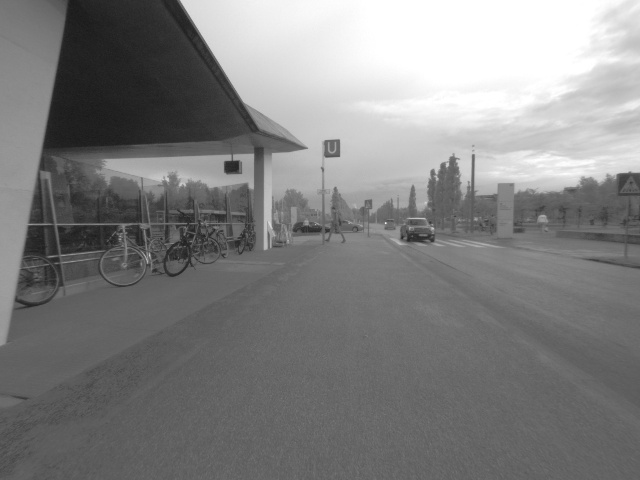}
		\vspace*{-6.8mm}
		\caption*{\tiny{seq 46}}
	\end{subfigure}  
	\begin{subfigure}[b]{0.15\textwidth}
		\includegraphics[width=\textwidth]{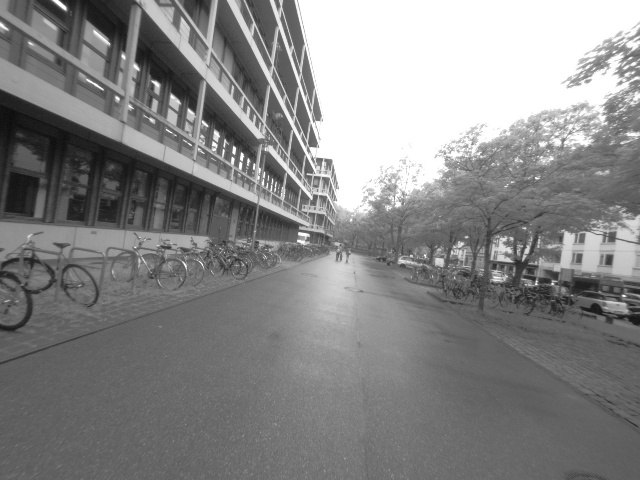}
		\vspace*{-6.8mm}
		\caption*{\tiny{seq 47}}
	\end{subfigure}
	\begin{subfigure}[b]{0.15\textwidth}
		\includegraphics[width=\textwidth]{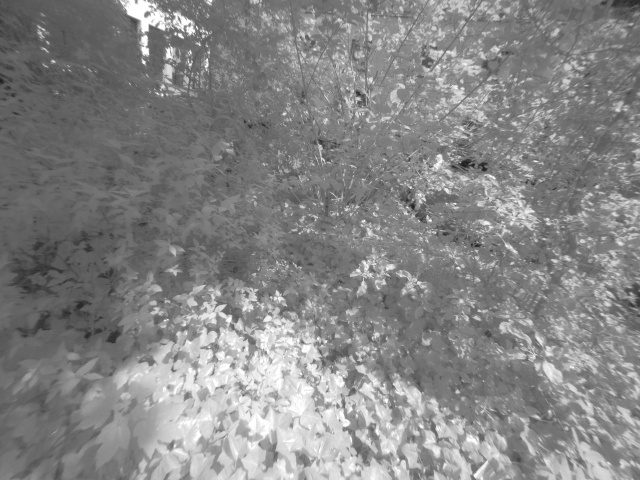}
		\vspace*{-6.8mm}
		\caption*{\tiny{seq 31}}
	\end{subfigure}
	\caption{Sample sequences from the TUM Mono VO Dataset, on which ORB-SLAM has the largest motion bias~\cite{tummono}. The first 5 images
		show scenarios where motion bias can happen. The end part of \emph{sequence 31}
		is shown in the last image. Such high frequency textures (leaves) are challenging
		for initialization when running backward.}
	\label{fig:biaspickframes}
	\vspace{-2.0em}
\end{figure}

To verify this, we check the sequences on which ORB-SLAM performs significantly 
better running backward and show them in the first 5 subfigures in 
Fig.~\ref{fig:biaspickframes}. It can be seen that all of them fulfill our 
description above. We also check the two counter examples, \emph{sequence 31} and 
\emph{44}, on which ORB-SLAM performs better running 
forward. At the end part of \emph{sequence 31} there is a large amount of high 
frequency textures (leaves) as shown in the last image in 
Fig.~\ref{fig:biaspickframes}, which makes ORB-SLAM not able to initialize or 
fail directly after the initialization when running backward. In \emph{sequence 
44}, interestingly, the camera is moving most of the time backward, which in 
fact verifies our conclusion.

\begin{figure}
\vspace{1.0em}
	\centering
	\includegraphics[width=0.8\linewidth]{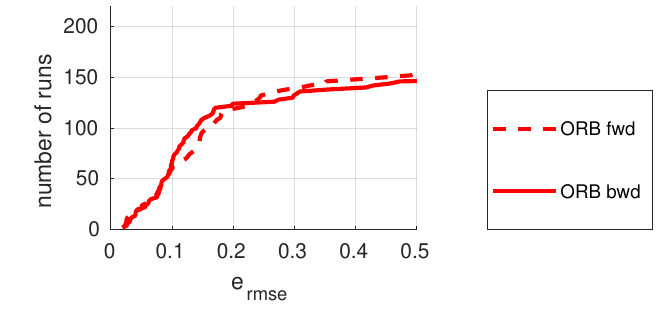}
	\vspace*{-2mm}
	\caption{Performance of ORB-SLAM on EuRoC MAV Dataset running forward and backward.}
	\label{fig:mavresult}
	\vspace{-1.5em}
\end{figure}

We also run ORB-SLAM on the EuRoC MAV Dataset, 
where the sequences are captured in relatively closed indoor environment and 
the camera motion is rather diverse without any clear pattern. We assume 
ORB-SLAM should deliver similar results running forward and backward. The 
result is shown in Fig.~\ref{fig:mavresult} and it verifies our assumption.

The analysis above does not explain why DSO performs consistently running 
forward and backward. We claim the performance gain of DSO mainly comes 
from its implementation and feature-based or semi-direct methods can be 
improved taking into consideration the following issues:

\textbf{a) Depth representation}. Instead of using depth directly like 
ORB-SLAM, DSO uses an inverse depth parametrization that affects the validity 
range of linearization and can better cope with distant 
features~\cite{civera2008inverse}. We thus claim that the distant points, which 
are poorly initialized from the image center, have less impact on DSO. 

\textbf{b) Point sampling strategy}. DSO samples points evenly across the 
entire image, which can be beneficial to avoid selecting many points from 
locations that only give poor initializations (\eg, image center). 


\textbf{c) Point management}. In ORB-SLAM, features extracted from a new frame 
will be added into the system, if they can match those features that are 
already in the window but haven't been matched before. If all these features 
gather together at the image center, they will be added with inaccurate 
depth estimations. In contrary, DSO only samples candidate points from the new 
frame but does not add them to the system immediately. The depth estimations of 
these points keep being refined (outliers are removed) before they are 
activated and added. Moreover, points are only selected to be activated
if they can keep the uniform spatial distribution of all activated points. All 
these strategies prevent problematic points from being added into the system.

\begin{figure*}
\vspace{1.0em}
	\centering
	\includegraphics[width=\linewidth]{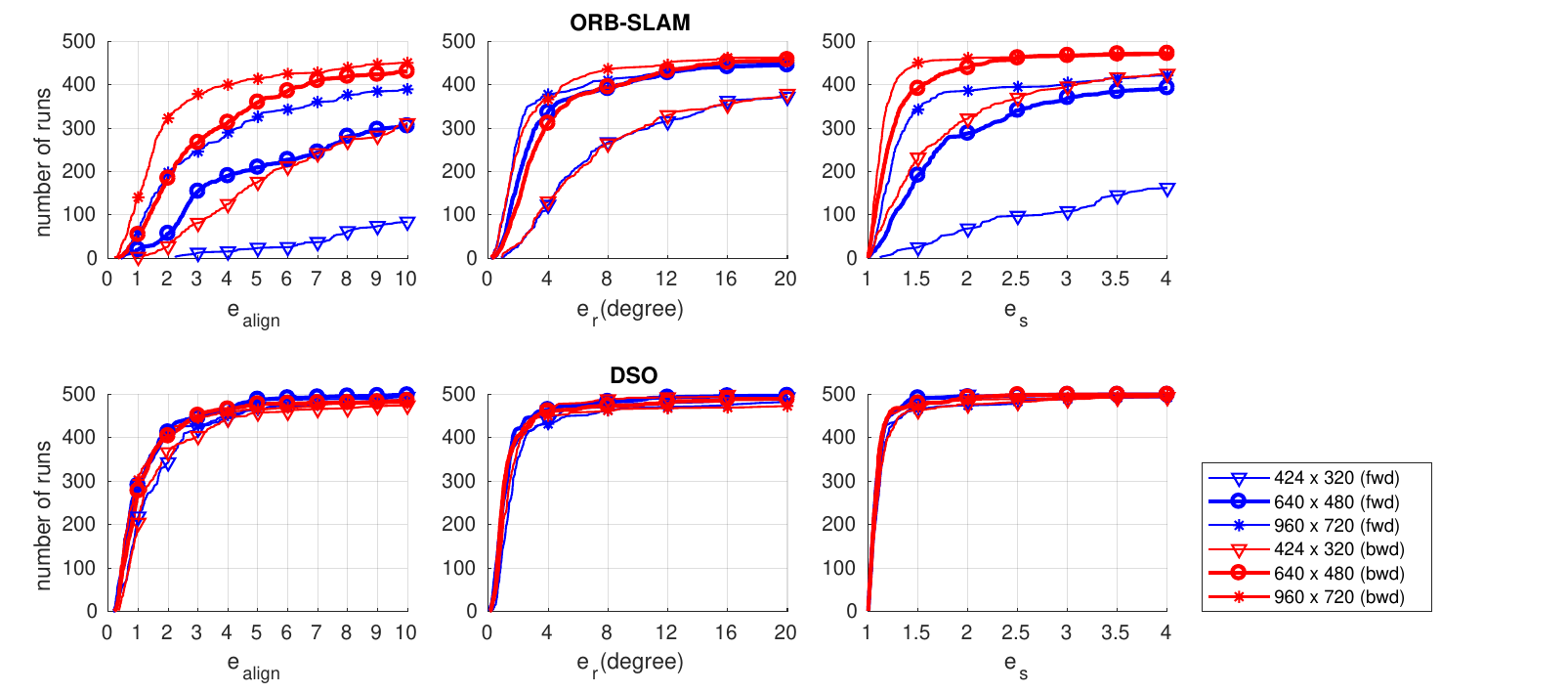}
	\vspace*{-4.8mm}
	\caption{Performance differences of ORB-SLAM (top) and DSO (down) on the 
		TUM Mono VO Dataset due to motion bias at different image resolutions. While DSO delivers
		similar results under different settings, ORB-SLAM performs consistently
		better when running backward and the performance gaps increase with reduced
		image resolutions.}
	\label{fig:biasresolution}
	\vspace{-1.0em}
\end{figure*}

\textbf{d) Discretization artifacts}. In direct methods, the depth of a newly 
observed point is initialized by searching for its correspondence in the 
reference frame along the epipolar line using sub-pixel accuracy. In 
feature-based methods, however, a new feature is extracted and matched to a 
previously observed feature with both of them at discretized image locations.
Thus feature-based methods suffer one time more from pixel discretization 
artifact. The effect becomes more severe when 
matching those distant features emerging from the image center running forward. 
To verify our analysis, we first perform the experiment in 
Fig.~\ref{fig:biasresolution} where we run DSO and ORB-SLAM forward and 
backward on sequences sampled to different resolutions. The performance of DSO 
drops a little on low resolution sequences, but overall it is robust
to such artifact. In contrast, the performance gaps of ORB-SLAM between running 
forward and backward increase significantly with reduced resolutions (thus 
severer discretization artifact). 

\begin{figure}
	\centering
	\includegraphics[width=\linewidth]{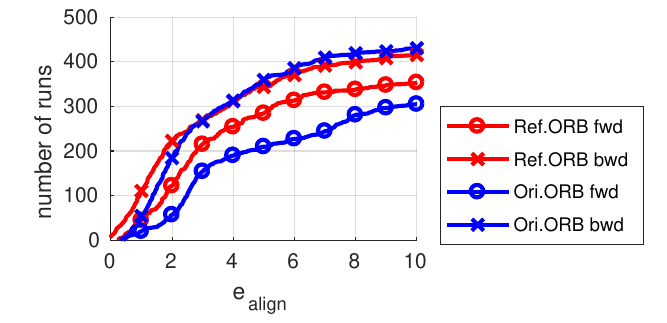}
	\vspace*{-4.8mm}
	\caption{Performance comparison between the original ORB-SLAM and our refined version
	on the full TUM Mono VO Dataset. With sub-pixel accuracy refinement of feature matching,
	ORB-SLAM performs better running forward and similarly running backward.}
	\label{fig:refined}
	\vspace{-0.8em}
\end{figure}

In our second experiment, we adopt a sparse optical flow algorithm to refine 
the feature matching step of ORB-SLAM to achieve sub-pixel precision. We use 
the iterative Lucas-Kanade method implemented in OpenCV and run the refined 
ORB-SLAM on the first 5 sequences shown in Fig.~\ref{fig:biaspickframes}.
The result is shown in Table~\ref{table:refined}. ORB-SLAM performs similarly 
running backward as before but much better (more than 50\% on average) running 
forward, which supports our analysis. For reference we also show the results on 
all the sequences in Fig~\ref{fig:refined}. 

\begin{table}
\vspace{1.0em}
	\small
	\begin{center}
		\begin{tabular}{|c|c|c||c|c|}
			\hline
			& \multicolumn{2}{c||}{\textbf{$e_{align}^{fwd}$}} & \multicolumn{2}{c|}{\textbf{$e_{align}^{bwd}$}}\\
			\hline
			Seq. & \multicolumn{1}{c}{Original} & \multicolumn{1}{c||}{Refined} & \multicolumn{1}{c}{Original} & \multicolumn{1}{c|}{Refined}\\
			\hline
			\hline
			17   & \multicolumn{1}{c}{12.29} & \multicolumn{1}{c||}{\textbf{3.05}}  & \multicolumn{1}{c}{1.50} & \multicolumn{1}{c|}{\textbf{1.33}}\\
			23   & \multicolumn{1}{c}{10.33} & \multicolumn{1}{c||}{\textbf{5.52}}  & \multicolumn{1}{c}{3.18} & \multicolumn{1}{c|}{\textbf{1.94}}\\
			29   & \multicolumn{1}{c}{21.84} & \multicolumn{1}{c||}{\textbf{10.52}} & \multicolumn{1}{c}{\textbf{2.24}} & \multicolumn{1}{c|}{2.58}\\
			46   & \multicolumn{1}{c}{27.18} & \multicolumn{1}{c||}{\textbf{14.89}} & \multicolumn{1}{c}{\textbf{5.10}} & \multicolumn{1}{c|}{5.27}\\
			47   & \multicolumn{1}{c}{20.57} & \multicolumn{1}{c||}{\textbf{10.85}} & \multicolumn{1}{c}{\textbf{4.58}} & \multicolumn{1}{c|}{5.80}\\
			\hline
			\hline
			mean & \multicolumn{1}{c}{18.44} & \multicolumn{1}{c||}{\textbf{8.97}}  & \multicolumn{1}{c}{\textbf{3.32}} & \multicolumn{1}{c|}{3.38} \\
			\hline
		\end{tabular}
	\end{center}
	\caption{Results of original ORB-SLAM and our refined ORB-SLAM. We run both methods 10 times on each of the selected sequences
		from Fig. \ref{fig:biaspickframes}. The data of original ORB-SLAM is 
		obtained from\cite{tummono}. Sub-pixel refinement significantly 
		improved the performance of forward run, while did not help so much 
		with the backward run, which verifies our conclusion.}
\label{table:refined}
\vspace{-1.5em}
\end{table}

\subsection{Rolling Shutter Effect}
\label{sec:rolling_shutter}

\begin{figure}
	\centering
	\captionsetup[sub]{font=small, justification=centering}
	\begin{subfigure}[b]{0.8\linewidth}
		\includegraphics[width=\linewidth]{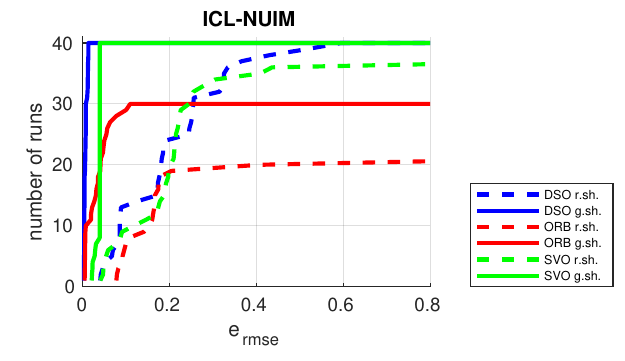}
		\caption{Performances of the selected methods.}
		\label{fig:iclresults_full}
	\end{subfigure}
	\begin{subfigure}[b]{0.48\linewidth}
	\vspace*{2.0mm}
		\includegraphics[width=\textwidth]{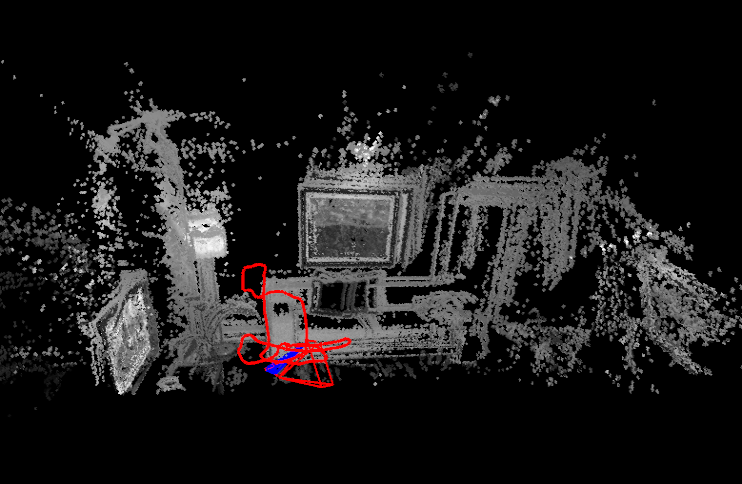}
		\caption{DSO, \emph{lr kt0} global shutter.}
		\label{fig:icl_global}
	\end{subfigure}	
	\begin{subfigure}[b]{0.48\linewidth}
	\vspace*{2.0mm}
		\includegraphics[width=\linewidth]{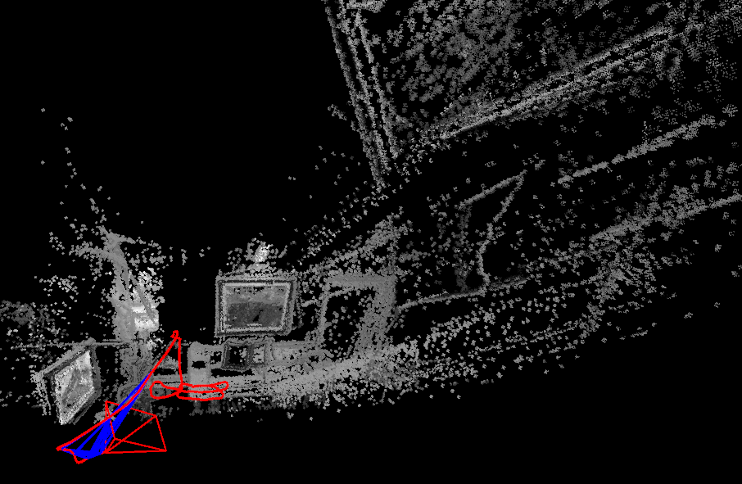}
		\caption{DSO, \emph{lr kt0} rolling shutter.}
		\label{fig:icl_rolling}
	\end{subfigure}	
	\caption{Results on the extended ICL-NUIM Dataset with original global shutter
		setting and the simulated rolling shutter setting. }
	\label{fig:iclresults}
	\vspace{-1.5em}
\end{figure}

\begin{figure}
\vspace{0.2em}
	\centering
	\captionsetup[sub]{font=small, justification=centering}
	\begin{subfigure}[b]{0.48\linewidth}
		\includegraphics[width=\textwidth]{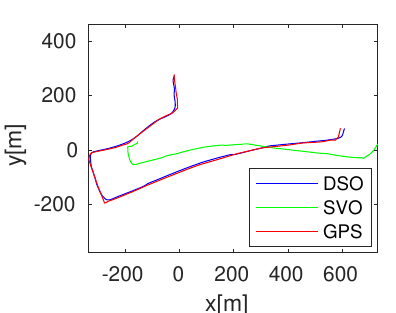}
		\caption{Frame 27001-33000}
		\label{fig:csresults_sub1}
	\end{subfigure}	
	\begin{subfigure}[b]{0.48\linewidth}
		\includegraphics[width=\linewidth]{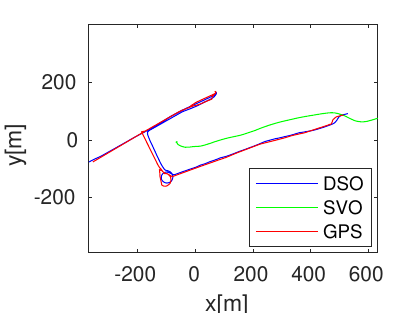}
		\caption{Frame 87001-93000}
		\label{fig:csresults_sub2}
	\end{subfigure}
	\caption{Estimated trajectories of the segments of the \emph{Frankfurt} sequence
	from the Cityscapes Dataset.
	The used frames are shown below each plot. 
		Estimated poses are aligned to GPS coordinates with 7D similarity transformation.
		Note that the provided GPS coordinates are not accurate.}
	\label{fig:csresults}
	\vspace{-1.5em}
\end{figure}

In the first experiment of this section, we run DSO, ORB-SLAM and SVO 10 times 
on each of the 4 \emph{Living Room} sequences of the ICL-NUIM Dataset, as well 
as on their simulated rolling shutter correspondences. Although currently this 
is the only dataset that provides global shutter and rolling shutter sequences 
at the same time, it is worth noting the simulated rolling shutter effect is 
relatively strong. The results are shown in Fig.~\ref{fig:iclresults_full}. All 
three methods are influenced by rolling shutter effect, yet the performance 
declines of DSO and SVO are apparently larger than ORB-SLAM. This result 
verifies that feature-based methods are more robust to the rolling shutter 
effect than direct methods. To show the influence of the rolling shutter effect 
on direct methods, we show examples of the reconstructed scene by DSO in 
Fig.~\ref{fig:icl_global} and Fig.~\ref{fig:icl_rolling}. It can be seen that 
the delivered reconstruction has very large scale drift on the rolling shutter 
sequence (the big structure in the background of Fig.~\ref{fig:icl_rolling} is 
the drifted reconstruction of the painting in the foreground).

Although SVO performs feature matching followed by BA for refining structures 
and poses, in the initial pose estimation for each frame it uses direct image 
alignment, thus does not use the correspondences estimated by feature matching. 
This explains its performance decline is larger than that of ORB-SLAM.
It is also worth mentioning, as can be seen in Fig.~\ref{fig:iclresults_full},
that the overall performances of DSO and SVO on this dataset significantly 
transcend the one of ORB-SLAM under both global and rolling shutter settings. 
The main reason is that the scenes in this dataset are indoor environments with 
low textured structures such as walls, floors and doors and thus are very 
challenging for corner-based feature extraction. Due to the fact the SVO is 
able to use image information on edges, it gains certain robustness on this 
dataset. As a result, the selected direct and semi-direct methods outperform 
the feature-based method even on rolling shutter images. 

While the results above coincide with our intuition, it sometimes can be 
misleading. One may easily draw the conclusion that on sequences with enough 
texture and captured using rolling shutter cameras, feature-based methods 
should be preferable than direct or semi-direct methods. This is not always the 
case. Recall that the rolling shutter effect in the extended ICL-NUIM Dataset 
is artificially simulated. On modern industrial level cameras, pixel read-out 
speeds are usually extremely fast such that the rolling shutter effect is to 
some extent neglectable for many applications. 
In the second experiment, we aim at comparisons on images with such realistic 
rolling shutter effect. As there is no such dataset that provides both real 
global shutter and rolling shutter sequences, we only compare the VO accuracies 
on realistic rolling shutter sequences. For this purpose, we use the 
\emph{Frankfurt} sequence of the Cityscapes Dataset and split it into
smaller segments (each with around 6000 frames). To our surprise, ORB-SLAM 
always fails on the selected segments: whenever the camera rotates strongly at 
street corners or large occlusion occurs due to moving vehicles, which has also 
been reported by other users of ORB-SLAM. We thus suspect the failures are not 
related to the rolling shutter effect. In Fig.~\ref{fig:csresults} we show the 
estimated camera trajectories of DSO and SVO. Although SVO suffers more from 
scale drift, both the direct and semi-direct methods are able to track on the 
entire selected segments.
The last thing to point out is, without a proper dataset, it is still difficult 
to analyze the exact influence of the rolling shutter effect on existing VO 
methods. 

%% file: Conclusion.tex
\section{Conclusions}
We present a thorough evaluation for state-of-the-art direct, semi-direct and 
feature-based methods on photometric calibration, motion bias and the rolling 
shutter effect, with the aim of providing practical inputs to the community for 
better applying existing methods and developing new VO and SLAM algorithms.
Our main conclusions are: 

\textbf{(1)} With photometric calibration, the performance of
direct methods gets improved significantly, while for semi-direct and 
feature-based methods, it depends on the used feature, the camera response 
function and the overall brightness of the scene. Ideally active camera control 
~\cite{zhang2017active} should be applied to deliver the feature extractor and 
matcher with images of good quality.
For direct methods, when photometric calibration information is not available, 
online calibration methods~\cite{bergmann2018online} should be used. 

\textbf{(2)} Compared to direct methods, feature-based methods have a 
relatively large performance bias when running forward and backward. Possible 
reasons are discussed: depth representation, point selection and management, 
discretization artifact. When adopting existing feature-based methods for 
applications like autonomous driving, more effort should be taken to address 
the motion bias.

\textbf{(3)} Direct and semi-direct methods are more sensitive to the rolling 
shutter effect. But when the rolling shutter effect is not strong, or the 
environment is low textured, the rolling shutter effect might not be the 
deciding factor on performance anymore. When the pixel readout speed is fast 
enough, even direct methods can deliver satisfying results. Besides, a specific 
dataset is needed for getting a better understanding on the rolling shutter 
effect. 

\textbf{(4)} The used feature-based methods are more sensitive to pixel 
discretization artifact. When possible, images with higher resolutions are 
preferable. Moreover, sub-pixel accuracy refinement on feature extraction and
matching can boost their performance, which is verified by our sub-pixel 
refined version of ORB-SLAM. \\


\noindent{\textbf{Acknowledgement}}. We would like to thank Jakob Engel and 
J{\"o}rg St{\"u}ckler for the insightful discussions. We also would like to 
thank Zichao Zhang to recommend the settings of SVO 2.0 for the evaluation on 
the TUM Mono VO dataset.